\documentclass[letterpaper,10pt,conference]{ieeeconf}

\IEEEoverridecommandlockouts
\usepackage{amsmath,amsfonts,amssymb}
\usepackage{algorithmic}
\usepackage{algorithm}
\usepackage{array}
\usepackage[caption=false,font=footnotesize,labelfont=rm,textfont=rm]{subfig}
\usepackage{textcomp}
\usepackage{stfloats}
\usepackage{url}
\usepackage{verbatim}
\usepackage{graphicx}
\usepackage{cite}
\usepackage{multirow}
\usepackage{booktabs}


\begin{document}

\title{\LARGE \bf SOR-Nav: Search or Relocate? Context-Gated Exploration and Cross-Region Relocation for Object Navigation}

	\author{Yuan Ji,
		Zirui Li,
		Yuxin Cai,
		Shuge Wu,
		Boon Siew Han,
		and Chen Lv*%
		\thanks{Yuan Ji, Zirui Li, Yuxin Cai, Shuge Wu, and Chen Lv are with the School of Mechanical and Aerospace Engineering, Nanyang Technological University, 639798, Singapore (e-mail: \{yuan.ji, zirui.li, shuge.wu\}@ntu.edu.sg; caiy0039@e.ntu.edu.sg; lyuchen@ntu.edu.sg).}
		\thanks{Boon Siew Han is with Schaeffler Hub for Advance REsearch(SHARE) at NTU, Singapore (e-mail: hanbon@schaeffler.com).}
		\thanks{This research is supported by the RIE2025 Industry Alignment Fund -- Industry Collaboration Projects (IAF-ICP)Grant No. I2501E0041), administered by A*STAR, as well as supported by Schaeffler (Singapore) PTE. LTD. and NTU Singapore through Schaeffler-NTU Corporate Lab: Intelligent Mechatronics Hub.}
	}

\maketitle
\thispagestyle{empty}
\pagestyle{empty}

\begin{abstract}
	Object navigation requires an embodied agent to find an object in an unseen environment under partial observability and a limited motion budget.
	Existing methods primarily optimize where the robot should go next by ranking candidate destinations. In contrast to these methods, we present SOR-Nav, a hierarchical navigation system that explicitly arbitrates between continuing to explore the current context and abandoning it for a more promising reachable region.
	First, an autonomous semantic exploration system is built that accumulates persistent 3D object clusters and organizes reachable frontiers into a cluster decision graph to provide an efficient search abstraction.
	Then, SOR-Nav uses a context-gated LLM-driven object-search supervisor to evaluate the suitability of the current search context and decide whether to continue exploration or perform cross-region relocation to another reachable frontier cluster.
	Across the complete, unfiltered validation sets of HM3D-v1, HM3D-v2, and MP3D, SOR-Nav achieves the strongest reported Success Rate (SR) and Success weighted by Path Length (SPL) on all three benchmarks. On MP3D in particular, it more than doubles the previous best SPL from 18.1\% to 38.5\% while increasing SR from 50.7\% to 61.8\%.
	Nested HM3D-v2 ablations validate the proposed decision structure, while a continuous three-target physical deployment demonstrates persistent ObjectNav operation in real-world scenarios.
\end{abstract}


\section{Introduction}

Object navigation (ObjectNav) requires an embodied agent to locate an instance of a requested object category in an unseen environment and stop near it.
Although the goal is specified by only a category name, successful execution couples semantic perception, persistent spatial memory, exploration, target verification, and motion planning.
The target may lie outside the initial field of view or several rooms away, while every detour consumes a finite action and path budget.
ObjectNav is therefore not simply an object-recognition problem: it is a long-horizon decision problem under partial observability.

Recent zero-shot systems use language and vision foundation models to inject object--room commonsense or image--language similarity into spatial search~\cite{esc,l3mvn,vlfm,openfmnav,sgnav,voronav}.
Their common question is prospective: which visible direction, frontier, room, or candidate region is most likely to lead to the target?
This semantic guidance is valuable when observations are informative, but it can be brittle when evidence is sparse, occluded, or misleading.
A locally plausible candidate may keep the robot within a semantically unsuitable context, and repeatedly rescoring nearby frontiers may consume the budget without producing a meaningful region change.
Robustness mechanisms improve recovery, candidate verification, and system decomposition~\cite{trihelper,apexnav,intentnav,sysnav}, yet they do not make search persistence itself the central decision.

We address the complementary question: when no target has been confirmed, is the current search context still worth exploring?
This distinction matters because continuing and leaving are not symmetric choices.
Uncertainty alone should not force a costly cross-region relocation, but clear evidence that the current context is unsuitable should allow the system to move to a more promising reachable region.
Candidate attractiveness should therefore determine the destination only after the evidence provides a reason to leave.
Here, \emph{cross-region relocation} means terminating exploration in the current search context and navigating to another graph-reachable frontier cluster, rather than merely selecting a different local frontier.

We propose SOR-Nav, a hierarchical ObjectNav system comprising an autonomous semantic exploration system and a context-gated, LLM-driven object-search supervisor.
The exploration system accumulates semantic observations as persistent 3D object clusters and organizes reachable frontiers into a cluster decision graph.
When no reliable target is available, the supervisor decides whether to continue receding-horizon exploration from the current context or relocate to another graph-reachable frontier cluster.
Semantic context is used to judge the suitability of the current and candidate regions, while graph structure preserves reachability, coverage, and travel cost.
The resulting decision remains grounded in a shared waypoint executor and is revised whenever new observations or mission outcomes arrive.

The main contributions are:
\begin{itemize}
	\item We introduce a search-or-relocate formulation for object navigation tasks. Unlike existing methods that primarily decide where the robot should go next, SOR-Nav instead asks whether the current search context should be abandoned and considers alternative destinations only after cross-region relocation is justified.
	\item To operationalize this formulation, we develop a hierarchical ObjectNav decision system comprising an autonomous semantic exploration system and a context-gated, LLM-driven supervisor. Persistent 3D object clusters provide semantic context, while a traversability-preserving cluster decision graph supplies reachable frontier clusters, enabling the supervisor to coordinate receding-horizon exploration and cross-region relocation through executable goals.
	\item Across the complete evaluations of the standard ObjectNav benchmarks---HM3D-v1, HM3D-v2, and MP3D---SOR-Nav consistently achieves the strongest performance, especially on MP3D, where it more than doubles the best reported SPL from 18.1\% to 38.5\% while increasing SR from 50.7\% to 61.8\%. A continuous three-target real-world experiment further demonstrates persistent ObjectNav operation without resetting the map.
\end{itemize}

\section{Related Work}

\subsection{Object Navigation}

End-to-end ObjectNav policies learn observation-to-action mappings through large-scale reinforcement learning, human demonstrations, or their combination~\cite{ddppo,habitatweb,pirlnav}.
Modular systems instead expose semantic perception, mapping, goal selection, planning, and control as separate components.
SemExp accumulates category observations in a semantic map, and PONI learns category-conditioned spatial potentials~\cite{semexp,poni}; these systems established the map-mediated pipeline used by many later zero-shot methods.
The distinction is consequential for long-duration search.
An end-to-end policy must encode recognition confidence, spatial memory, recovery, and stopping behavior in a single recurrent decision process, whereas a modular system can preserve each source of evidence beyond the observation that produced it.
SOR-Nav adopts the latter organization because a search-or-relocate decision must compare the current search context with regions observed at different times and hand any selected region to an independently executable navigation stack.

Foundation-model approaches primarily change how spatial candidates are valued.
ESC and L3MVN use object--room commonsense to guide frontier selection, while VLFM and OpenFMNav project language-aligned visual relevance into spatial value maps~\cite{esc,l3mvn,vlfm,openfmnav}.
SG-Nav prompts an LLM with a hierarchical scene graph, and VoroNav supplies topological waypoints rather than dense frontier pixels~\cite{sgnav,voronav}.
TriHelper adds recovery and target verification, ApexNav switches between semantic and geometric exploration and orders promising frontiers, and IntentNav retains candidate evidence and trajectory history in a shared BEV decision space~\cite{trihelper,apexnav,intentnav}.
These methods improve what the agent should approach; SOR-Nav focuses on the orthogonal persistence decision of whether the present search context should be retained or abandoned.
This decision also differs from collision recovery or short-horizon action correction: the robot may be navigating successfully at the control level while searching the wrong spatial context at the task level.

\subsection{Structured Representations and Navigation Systems}

Dense occupancy, BEV, semantic, and value maps preserve metric detail and observation history~\cite{semexp,poni,l3mvn,esc,vlfm,openfmnav,instructnav}, but a decision layer often reduces them to independently scored candidates.
Structured representations expose longer-range relations: VoroNav uses a reduced Voronoi graph, SG-Nav organizes objects and rooms hierarchically, and SysNav reasons at room level while reserving fine-grained coverage for classical navigation~\cite{voronav,sgnav,sysnav}.
IntentNav similarly separates candidate-level decision making from low-level execution, but represents history directly in BEV~\cite{intentnav}.
Across these designs, abstraction serves two purposes: it reduces the input presented to a high-level reasoner and constrains its output to locations that a planner can execute.
Both are important for SOR-Nav, since an unconstrained semantic suggestion may be plausible in language space yet absent or unreachable in the partially mapped environment.

SOR-Nav uses structure for a different comparison.
It first groups graph-connected frontiers into reachable exploration regions and only then attaches persistent semantic context.
Consequently, semantic compatibility does not overwrite connectivity or travel cost, and nearby frontiers separated by a wall are not treated as the same region.
The supervisor can therefore compare continued exploration in the current search context with an executable cross-region relocation.
This emphasis on region departure complements prior work on semantic attraction and follows the system-level separation of reasoning, planning, and control advocated by SysNav~\cite{sysnav}.

\section{Method}

\subsection{System Overview}

\begin{figure*}[t]
	\centering
	\includegraphics[width=0.8\textwidth]{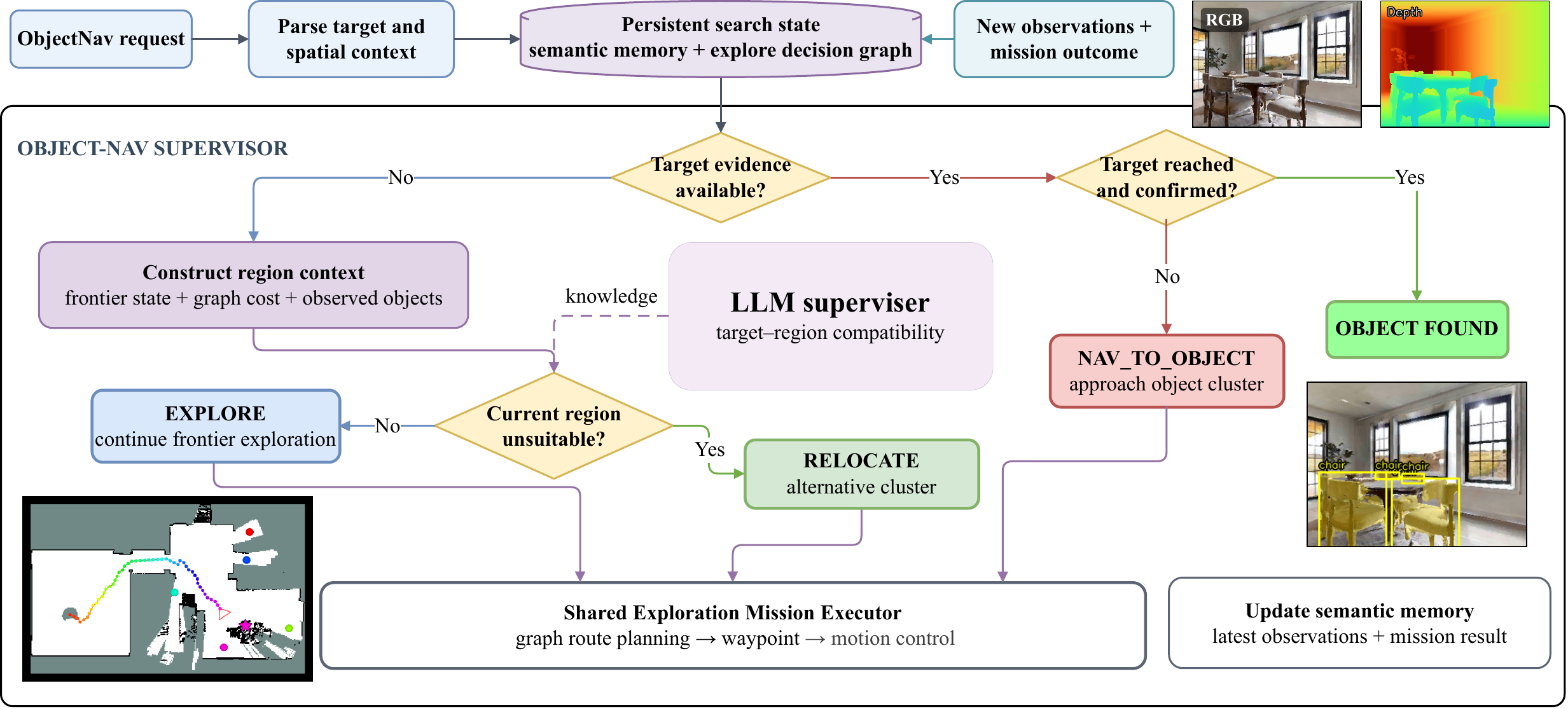}
	\caption{Overview of the SOR-Nav object-search workflow.
		RGB and depth observations update persistent 3D object clusters and a cluster decision graph in the autonomous semantic exploration system.
		The LLM-driven supervisor prioritizes confirmed target evidence; otherwise, its context gate determines whether to continue exploration or relocate to a graph-reachable frontier cluster.
		All outcomes share the same mission executor, and subsequent observations and mission results close the loop.}
	\label{fig:expnav_overview}
\end{figure*}

Given a target category and sequential RGB-D observations, SOR-Nav's autonomous semantic exploration system maintains persistent 3D object clusters and a cluster decision graph of traversable space and reachable frontiers.
As shown in Fig.~\ref{fig:expnav_overview}, a context-gated, LLM-driven object-search supervisor reads these states and chooses one of three mission-level actions: approach a reliable target candidate, continue exploring the current context, or relocate to another reachable frontier cluster.

The ordering of these decisions is important.
Reliable target evidence preempts exploration and triggers target-conditioned navigation.
When no target is available, the supervisor evaluates whether the current semantic context remains compatible with the request.
Uncertain or incomplete evidence preserves exploration; only a clearly unsuitable context opens the relocation gate and enables comparison of alternative clusters.
This asymmetric rule prevents every weak semantic preference from causing a costly cross-region relocation.

All decisions use a shared mission executor.
Continued exploration sends the next frontier from a receding-horizon route, relocation sends an executable waypoint associated with the selected cluster, and target navigation sends the center of the confirmed object cluster.
Mission outcomes and new observations update both memories before the next decision.
The supervisor therefore regulates the spatial scale and purpose of navigation without replacing the underlying mapping, path planning, or motion control modules.

The modules operate at different temporal scales and exchange only persistent state and waypoint-level missions.
Semantic fusion and graph construction follow incoming observations, the supervisor runs when its decision interval expires or the active mission terminates, and the motion stack continuously executes the current waypoint.
It also makes mission switching explicit: relocation changes the active search context but does not indicate task completion, while a target approach remains provisional until the target gate validates the terminal observation.

\subsection{Autonomous Semantic Exploration System}

Semantic labels are projected with depth into a common 3D frame and fused in a voxel map.
Connected voxels with the same label form an object cluster $o_j^t=(\ell_j,N_j,\mathbf{p}_j,\mathcal{B}_j)$, represented by its category, voxel support, center, and spatial bounds.
Repeated views increase spatial support instead of creating independent detections, while disconnected instances remain separate clusters.
This persistent representation suppresses isolated frame-level noise, allows target evidence to survive viewpoint changes, and supplies contextual objects even after they leave the camera view.

In parallel, the explorer constructs a global frontier decision graph
\begin{equation}
	\mathcal{G}_{F}^{t}
	=
	\left(
	\{v_{r}^{t},v_{h}\}\cup\mathcal{V}_{F}^{t},
	\mathcal{E}_{F}^{t}
	\right),
	\label{eq:frontier_decision_graph}
\end{equation}
containing the robot node $v_r^t$, graph-grounded frontier nodes $\mathcal{V}_F^t$, and an optional route terminal $v_h$ at the initial pose.
Each edge stores an executable path on the topological map and its traversal cost $w_{ij}^t$.
The cost can combine graph-path length with same-floor endpoint displacement; the Habitat experiments use path length alone, so decisions reflect traversability rather than Euclidean proximity.

Individual frontiers may appear, split, or disappear as free space is revealed.
To expose more stable ObjectNav decision units, SOR-Nav groups frontier nodes by low-cost graph connectivity:
\begin{equation}
	\begin{aligned}
		v_i\sim v_j
		 & \Longleftrightarrow
		\exists\text{ a frontier-edge path }i\rightsquigarrow j
		\text{ with }w_e^t\leq\rho_c, \\
		\mathcal{G}_{C}^{t}
		 & =
		\left(
		\{v_r^t\}\cup
		\{c(C):C\in\mathcal{V}_{F}^{t}/\!\sim\},
		\mathcal{E}_{C}^{t}
		\right).
	\end{aligned}
	\label{eq:cluster_decision_graph}
\end{equation}
This graph-based construction avoids merging frontiers that are close in image or metric space but separated by walls or long detours.
A cluster node is placed at the mean position of its members; robot-to-cluster and inter-cluster edges reuse the least-cost paths already stored in $\mathcal{G}_F^t$.
Specifically,
\begin{equation}
	\begin{aligned}
		w_{r,k}^{t}
		 & =\min_{v_i\in C_k^t}w_{r,i}^{t}, \\
		w_{k,l}^{t}
		 & =\min_{\substack{v_i\in C_k^t    \\v_j\in C_l^t}}w_{i,j}^{t}.
	\end{aligned}
	\label{eq:cluster_edge_costs}
\end{equation}
The frontier endpoint attaining $w_{r,k}^{t}$ becomes the cluster's executable relocation waypoint.
Because these values reuse paths already computed for the frontier graph, the abstraction adds no new global path queries.
We use $\rho_c=25.0$ graph-cost units.

Before semantic reasoning, clusters are geometrically pre-ranked by
\begin{equation}
	U_{\mathrm{geo}}(C_k^t)
	=
	\frac{n_k^t(1+\eta r_k^t)}{1+d_k^t},
	\label{eq:cluster_geometry_utility}
\end{equation}
where $n_k^t$ is the number of member frontiers, $r_k^t$ is their spatial spread, and $d_k^t$ is the robot-to-cluster graph distance.
With $\eta=0.1$, the score favors regions that expose more boundary structure while penalizing relocation cost.
Only the top $K=10$ clusters are passed to the supervisor.
For each candidate, its semantic context and the robot-local context are
\begin{equation}
	\begin{aligned}
		\mathcal{S}_k^t
		 & =\{(\ell_j,N_j,d(o_j^t,C_k^t)):N_j\geq N_{\min},\ d(o_j^t,C_k^t)\leq R_s\}, \\
		\mathcal{S}_r^t
		 & =\{(\ell_j,N_j):\|\mathbf{p}_j-\mathbf{p}_r^t\|_{xy}\leq R_r\},
	\end{aligned}
	\label{eq:semantic_contexts}
\end{equation}
where $d(o_j^t,C_k^t)$ is the minimum horizontal distance from object center $\mathbf{p}_j$ to a member frontier.
We use $R_s=R_r=10$\,m and discard object clusters with fewer than $N_{\min}$ voxels.
Attaching semantics after spatial clustering keeps target relevance separate from reachability and coverage.

When exploration continues, the current frontier graph also supplies a global visiting order.
For frontier permutation $\boldsymbol{\pi}$, SOR-Nav approximately solves
\begin{equation}
	\boldsymbol{\pi}_t^\star
	\in
	\arg\min_{\boldsymbol{\pi}}
	\left[
	w_{r,\pi_1}^{t}
	+
	\sum_{q=1}^{M_t-1}w_{\pi_q,\pi_{q+1}}^{t}
	+
	\widetilde{w}_{\pi_{M_t},h}^{t}
	\right],
	\label{eq:frontier_tsp}
\end{equation}
where $\widetilde{w}_{i,h}^{t}$ is zero for the open route used in Habitat and otherwise returns the robot to its initial pose.
Only the first frontier is executed before observations update the graph and the route is solved again.
A nearby frontier replaces that first choice only when it lies within 5\,m and the planned first leg is more than twice as long.
Thus, TSP provides global ordering while receding-horizon execution remains responsive to newly revealed space.

\subsection{Context-Gated LLM-Driven Supervisor}

At each supervisory update, the requested category $g$, current semantic context $\mathcal{S}_r^t$, and candidate tuples containing cluster id, graph distance, and semantic context are evaluated as
\begin{equation}
	(a_t,k_t)
	=
	\Pi_{\mathrm{sup}}
	\left(
	g,\mathcal{S}_r^t,
	\{(k,d_k^t,\mathcal{S}_k^t)\}_{C_k^t\in\widehat{\mathcal{C}}^t}
	\right),
	\label{eq:supervisor_policy}
\end{equation}
where $k_t=-1$ denotes continued exploration.
The key policy is asymmetric:
\begin{equation}
	a_t
	=
	\begin{cases}
		\texttt{continue\_explore},
		 & \neg\operatorname{Conflict}(g,\mathcal{S}_r^t), \\
		\texttt{relocate\_cluster}(k_t),
		 & \operatorname{Conflict}(g,\mathcal{S}_r^t).
	\end{cases}
	\label{eq:context_gate_rule}
\end{equation}
Here, $\operatorname{Conflict}$ means that accumulated local objects clearly contradict the expected context of the target.
Relevance, mixed evidence, missing observations, or uncertainty retain the default action.
Only after establishing this reason to leave does the supervisor compare candidate compatibility and graph distance to select a destination.
A distant cluster that merely looks more promising cannot by itself trigger relocation.

This ordering distinguishes context-gated cross-region relocation from conventional semantic frontier ranking.
The supervisor does not ask the semantic model to assign a scalar value to every frontier, and it does not replace the geometry-based candidate construction.
Instead, the model receives a compact snapshot of persistent object evidence and makes one bounded intervention in the navigation state machine.
If relocation is justified, the selected cluster id is resolved back to the stored graph path and waypoint; otherwise, the existing exploration route remains valid.

Algorithm~\ref{alg:object_search_supervisor} summarizes the complete arbitration order.
The target gate is evaluated on every update because newly fused evidence should interrupt an exploration mission promptly.
In contrast, region-level semantic reasoning is rate-limited and suppressed while a relocation is already in progress.
This prevents repeated LLM calls from replacing a still-valid relocation goal and gives the robot time to acquire observations from the selected context.

\begin{algorithm}[t]
	\caption{Context-Gated LLM-Driven Object-Search Supervisor}
	\label{alg:object_search_supervisor}
	\begin{algorithmic}[1]
		\STATE Update semantic memory $\mathcal{M}_t$ and graphs $\mathcal{G}_F^t,\mathcal{G}_C^t$
		\IF{target gate accepts a cluster of category $g$}
		\RETURN \texttt{NAV\_TO\_OBJECT}
		\ENDIF
		\IF{a relocation mission is active}
		\RETURN keep the current relocation waypoint
		\ENDIF
		\IF{the supervisor interval has not elapsed}
		\RETURN continue the current frontier mission
		\ENDIF
		\STATE Rank clusters by $U_{\mathrm{geo}}$ and retain $\widehat{\mathcal{C}}^t$
		\STATE Build $\mathcal{S}_r^t$ and $\{(k,d_k^t,\mathcal{S}_k^t)\}$
		\STATE $(a_t,k_t)\leftarrow\Pi_{\mathrm{sup}}(g,\mathcal{S}_r^t,\widehat{\mathcal{C}}^t)$
		\IF{$a_t$ is a valid relocation to $k_t$}
		\RETURN \texttt{RELOCATE} to the stored waypoint of $C_{k_t}^t$
		\ELSE
		\RETURN \texttt{EXPLORE} toward the first frontier in $\boldsymbol{\pi}_t^\star$
		\ENDIF
	\end{algorithmic}
\end{algorithm}

The current implementation instantiates $\Pi_{\mathrm{sup}}$ with an LLM under a constrained JSON action contract and a configurable decision interval.
The model receives label--support summaries rather than images or raw maps, limiting it to the intended region-level decision.
Invalid output, timeout, a disabled model, or an unknown cluster id falls back to continued exploration.
An active relocation is not interrupted by another exploration decision, and its successful completion immediately dispatches global exploration from the new region.
These rules make relocation a bounded supervisory intervention rather than a new local-search mode.

\subsection{Target Confirmation and Mission Execution}

Target evidence takes precedence over the context-gated search-or-relocate decision.
A target candidate becomes valid when its semantic cluster matches $g$ and contains at least $N_{\min}$ voxels.
Let $z_t$ indicate that such a cluster is present at update $t$.
The supervisor accumulates supporting observations $s_t$ and a consecutive miss counter $m_t$ as
\begin{equation}
	\begin{aligned}
		s_t & = s_{t-1}+z_t, \\
		m_t & =
		\begin{cases}
			0,         & z_t=1, \\
			m_{t-1}+1, & z_t=0.
		\end{cases}
	\end{aligned}
	\label{eq:target_evidence_update}
\end{equation}
Accumulation rejects a single-frame false positive, while the miss tolerance prevents brief occlusion during approach from immediately discarding a persistent candidate.
Final success requires
\begin{equation}
	s_t\geq N_{\mathrm{obs}},
	\qquad
	m_t\leq N_{\mathrm{miss}},
	\qquad
	b_t^{\mathrm{nav}}=1,
	\label{eq:target_confirmation}
\end{equation}
where $b_t^{\mathrm{nav}}$ denotes successful navigation verification.
We use $N_{\min}=10$, $N_{\mathrm{obs}}=3$, and $N_{\mathrm{miss}}=5$.

A valid visible candidate dispatches \texttt{NAV\_TO\_OBJECT}; otherwise, the supervisor applies~\eqref{eq:context_gate_rule}.
Target approach, frontier exploration, and cluster relocation all terminate at the same waypoint-level executor.
After reaching the target neighborhood, the adapter turns the camera toward the selected 3D cluster and requests a fresh observation before issuing \textsc{Stop}.
If the evidence has expired or the target can no longer be verified, the episode returns to the search supervisor instead of reporting success.
Likewise, completing a relocation immediately resumes global exploration from the new region rather than treating the relocation waypoint as a local goal.
After every mission, updated semantic evidence and graph state determine the next action, so neither a relocation waypoint nor an early target observation is treated as unconditional task completion.

\section{Experiments}

\subsection{Simulation Benchmark}

\subsubsection{Benchmarks and Evaluation Protocol}

We evaluate SOR-Nav on the complete validation episode lists used by our Habitat setup: 2,000 HM3D-v1 episodes, 1,000 HM3D-v2 episodes, and 2,195 MP3D episodes.
For each dataset, the recorded dataset indices are unique and span the full configured range.
No episode is removed based on floor difference, anticipated reachability, target category, or outcome; failures and timeouts therefore remain in the denominator.
Published baseline numbers in Table~\ref{tab:objectnav_benchmark} are taken from the simulation comparison reported by SysNav~\cite{sysnav}, whereas the SOR-Nav row is computed from our designated complete evaluation runs.
Because the cited systems differ in success-distance, target-instance, and evaluation-stack conventions, this table is an informative cross-paper comparison rather than a controlled reimplementation of every baseline.

\subsubsection{Implementation Details}

All three datasets use the same SOR-Nav mapping, exploration, target-verification, and navigation parameters.
Habitat provides synchronized RGB, depth, and semantic observations at $640\times480$ resolution with a $79^{\circ}$ horizontal field of view; the camera height is $0.88$~m and the simulated agent radius is $0.18$~m.
The action space consists of discrete forward motion, $30^{\circ}$ left and right rotations, and \textsc{Stop}; sliding is disabled and an episode is limited to 500 actions.
At episode start, the agent performs 12 scan actions to initialize its local observations, after which all motion is generated through the waypoint interface described in Sec.~III.
The map and object-search state are reset between episodes.
The adapter issues \textsc{Stop} only after the target gate reports navigation-verified evidence and the agent is within the configured 1.0~m target boundary.

\subsubsection{Metrics}

We report Success Rate (SR) and Success weighted by Path Length (SPL), the standard metrics used in Habitat ObjectNav.
SR measures the fraction of episodes in which the agent successfully reaches a target instance.
SPL penalizes successful episodes by the ratio between shortest path length and the executed path length.
We also report auxiliary statistics such as distance-to-goal and episode length when analyzing failure modes.

\subsubsection{Benchmark Results}

\begin{table*}[t]
	\centering
	\begin{minipage}{0.9\textwidth}
		\centering
		\caption{ObjectNav comparison based on the SysNav simulation table and SOR-Nav evaluation results.
			SR and SPL are reported in percentages.}
		\label{tab:objectnav_benchmark}
		\scriptsize
		\setlength{\tabcolsep}{1.0pt}
		\renewcommand{\arraystretch}{1.12}
		\begin{tabular*}{\linewidth}{@{\extracolsep{\fill}}lcrrrrrr@{}}
			\toprule
			\multirow{2}{*}{Method} & \multirow{2}{*}{ZS}
			& \multicolumn{2}{c}{HM3D-v1} & \multicolumn{2}{c}{HM3D-v2}
			& \multicolumn{2}{c}{MP3D} \\
			\cmidrule(lr){3-4}\cmidrule(lr){5-6}\cmidrule(lr){7-8}
			& & SR & SPL & SR & SPL & SR & SPL \\
			\midrule
			L3MVN~\cite{l3mvn}       & Yes & 50.4 & 23.1 & 36.3 & 15.7 & 34.9 & 14.5 \\
			ESC~\cite{esc}           & Yes & 39.2 & 22.3 & --   & --   & 28.7 & 11.2 \\
			VoroNav~\cite{voronav}   & Yes & 42.0 & 26.0 & --   & --   & --   & --   \\
			VLFM~\cite{vlfm}         & Yes & 52.5 & 30.4 & 63.6 & 32.5 & 36.4 & 17.5 \\
			SG-Nav~\cite{sgnav}      & Yes & 54.0 & 24.9 & 49.6 & 25.5 & 40.2 & 16.0 \\
			OpenFMNav~\cite{openfmnav} & Yes & 54.9 & 24.4 & --   & --   & 37.2 & 15.7 \\
			TriHelper~\cite{trihelper} & Yes & 56.5 & 25.3 & --   & --   & --   & --   \\
			InstructNav~\cite{instructnav} & Yes & -- & -- & 58.0 & 20.9 & -- & -- \\
			ApexNav~\cite{apexnav}   & Yes & 59.6 & 33.0 & 76.2 & 38.0 & 39.2 & 17.8 \\
			SysNav~\cite{sysnav}     & Yes & 63.7 & 30.5 & 80.8 & 37.2 & 50.7 & 18.1 \\
			\midrule
			SOR-Nav (ours)            & Yes & \textbf{66.1} & \textbf{36.9} & \textbf{87.6} & \textbf{52.7} & \textbf{61.8} & \textbf{38.5} \\
			\bottomrule
		\end{tabular*}
	\end{minipage}
\end{table*}

SOR-Nav achieves the highest reported SR and SPL in Table~\ref{tab:objectnav_benchmark} on all three datasets.
On HM3D-v1, it reaches 66.1\% SR and 36.9\% SPL over 2,000 episodes, improving over the strongest listed baseline by 2.4 SR points and 3.9 SPL points.
On HM3D-v2, it reaches 87.6\% SR and 52.7\% SPL over 1,000 episodes, corresponding to gains of 6.8 and 14.7 points, respectively.
The largest gap occurs on MP3D, where SOR-Nav obtains 61.8\% SR and 38.5\% SPL over 2,195 episodes, 11.1 SR points and 20.4 SPL points above the strongest listed result.
The consistently larger SPL margins indicate that the complete system not only finds more targets under our protocol but also limits path inflation on successful episodes.

To isolate the contributions of the object-search supervisor, we use two nested ablations on HM3D-v2.
The first disables context-gated cross-region relocation while retaining receding-horizon TSP ordering; the second additionally replaces TSP ordering with nearest-frontier execution.
All variants retain the same semantic inputs, target gate, navigation stack, and evaluation episodes.

\begin{table}[t]
	\centering
	\caption{Nested ablation study on HM3D-v2. SR, SPL, step-limit (SL), and timeout (TO) rates are percentages; $N$ is the number of evaluated episodes.}
	\label{tab:hm3d_v2_ablation}
	\scriptsize
	\setlength{\tabcolsep}{2.1pt}
	\renewcommand{\arraystretch}{1.08}
	\begin{tabular}{lrrrrr}
		\toprule
		Variant                     & $N$  & SR $\uparrow$ & SPL $\uparrow$ & SL $\downarrow$ & TO $\downarrow$ \\
		\midrule
		SOR-Nav (full)              & 1000 & 87.6          & 52.7           & 5.0             & 7.4             \\
		w/o relocation gate         & 1000 & 82.2          & 39.9           & 9.7             & 8.1             \\
		w/o relocation gate and TSP & 1000 & 76.4          & 29.2           & 16.3            & 7.3             \\
		\bottomrule
	\end{tabular}
\end{table}

Table~\ref{tab:hm3d_v2_ablation} shows that removing the context-gated cross-region relocation policy reduces SR by 5.4 points and SPL by 12.8 points, while increasing the step-limit rate from 5.0\% to 9.7\%.
Additionally replacing TSP-based frontier ordering with nearest-frontier execution causes a further 5.8-point SR drop and a 10.7-point SPL drop, with step-limit failures increasing to 16.3\%.
The timeout rate remains between 7.3\% and 8.1\% across the three variants; because timeout and step-limit exhaustion are competing terminal conditions, this small non-monotonic variation is not interpreted as an independent improvement.
Overall, the nested results indicate that context-gated cross-region relocation and global frontier ordering provide complementary benefits, primarily by reducing action-budget exhaustion and path inefficiency.

\subsection{Failure and Difficulty Analysis}

We further classify every unsuccessful v0 episode by its primary observable terminal condition, following the outcome-oriented analysis used by ApexNav~\cite{apexnav} while adapting the categories to the evidence recorded by our system.
An episode is labeled cross-floor when the vertical offset between the start and the closest annotated goal viewpoint exceeds 0.5~m.
Among the remaining failures, reaching 500 actions is labeled step-limit exhaustion.
Runs that instead reach the 120~s wall-clock budget are labeled timeout.
The single residual early termination is retained in the exported statistics but omitted from the primary-outcome plot.
The labels are assigned in this order and are therefore mutually exclusive; notably, cross-floor describes episode geometry rather than claiming a unique proximate failure mechanism.

\begin{figure}[t]
	\centering
	\includegraphics[width=0.95\columnwidth]{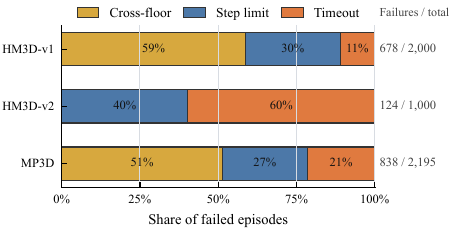}
	\caption{Primary outcomes among failed episodes in the complete v0 runs.
		Percentages are normalized within each dataset's failures; the right-side annotations report failed episodes over all evaluated episodes.}
	\label{fig:failure_composition}
\end{figure}

\begin{figure}[t]
	\centering
	\includegraphics[width=0.95\columnwidth]{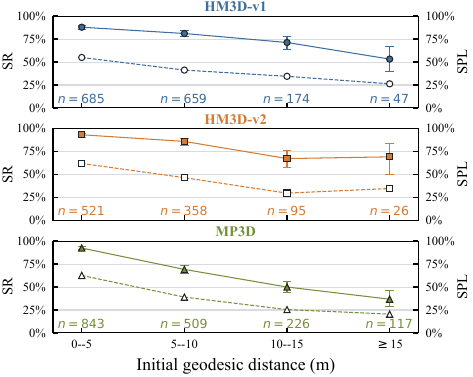}
	\caption{Same-floor navigation performance versus initial geodesic distance to the closest target.
		SR (solid lines, left axes) is shown with Wilson 95\% confidence intervals, while mean SPL (dashed lines with open markers) uses the right axes.
		Annotations give the number of episodes in each dataset--distance bin.}
	\label{fig:distance_analysis}
\end{figure}

\begin{figure*}[!t]
	\centering
	\includegraphics[width=0.9\textwidth]{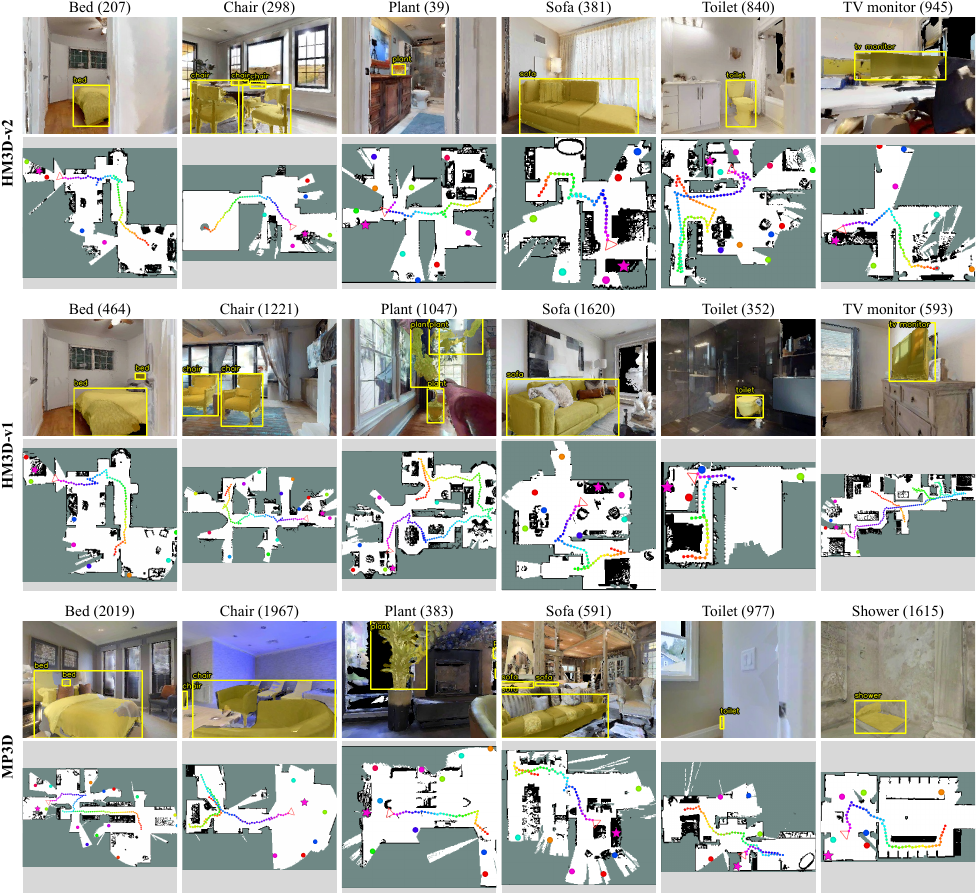}
	\caption{Paired qualitative results for 18 successful episodes across HM3D-v2, HM3D-v1, and MP3D, with six target categories shown per dataset.
		Within each episode panel, the goal-facing terminal RGB observation is shown directly above its corresponding explored occupancy map and executed trajectory.
		The yellow overlay marks the detected target category; the title reports the target and true dataset episode index.
		Portrait RViz maps with width-to-height ratio below 0.95 are rotated $90^{\circ}$ clockwise for compact display.}
	\label{fig:qualitative_results}
\end{figure*}

Figure~\ref{fig:failure_composition} shows that cross-floor geometry accounts for 58.7\% of the 678 HM3D-v1 failures and 51.3\% of the 838 MP3D failures.
After removing this factor, the same-floor SR is 82.1\% on HM3D-v1, 87.6\% on HM3D-v2, and 75.9\% on MP3D.
HM3D-v2 contains no cross-floor episodes under the 0.5~m definition; 50 of its 124 failures reach the step limit, while 74 reach the wall-clock timeout.
Across same-floor episodes, Fig.~\ref{fig:distance_analysis} reveals a second, independent difficulty axis: SR decreases from 87.9\% to 53.2\% on HM3D-v1 and from 92.4\% to 36.8\% on MP3D between the $0$--$5$~m and $\geq15$~m bins.
HM3D-v2 retains 69.2\% SR in its longest-distance bin, although the wider confidence interval reflects its smaller sample size ($n=26$).
SPL follows the same distance-dependent decline, falling from 0.55 to 0.26 on HM3D-v1, 0.62 to 0.35 on HM3D-v2, and 0.63 to 0.20 on MP3D.
These results identify multi-floor reasoning and long-horizon budget management as the two clearest remaining system-level limitations; they do not, by themselves, attribute individual failures to a particular supervisor component.

\subsection{Qualitative Results}

To inspect the terminal behavior of the current system, we re-executed 12 curated successful episodes from each dataset while recording periodic sensor views and RViz snapshots.
The HM3D sets cover all six benchmark goal categories in 12 distinct scenes; the MP3D set covers six categories and all 11 validation scenes.
All 36 re-executions remained successful.
These episodes were selected from historical successes to obtain diverse and readable visualizations, so this 36/36 result is a reproducibility check for the selected cases rather than an estimate of benchmark success.
Figure~\ref{fig:qualitative_results} presents six representative episodes from each dataset's curated set.

Figure~\ref{fig:qualitative_results} provides paired evidence that is obscured by aggregate metrics.
The 18 terminal views show that success corresponds to a visible instance of the commanded goal rather than only a distance threshold; after entering the success radius, the adapter aligns the camera with the selected 3D target before issuing \textsc{Stop}.
The corresponding occupancy views expose each episode's route geometry and preserve a continuous execution trace from the start region to the terminal target region.
Together, the panels demonstrate that the same perception--mapping--navigation interface operates across multiple target categories, scene collections, and spatial layouts.

\subsection{Real-World Demonstration}

We deployed SOR-Nav on an AgileX Scout Mini in a connected laboratory, exhibition, and office environment.
The robot carries an Insta360 panoramic RGB camera and a Livox MID-360 LiDAR.
Direct LiDAR odometry localizes the robot against a prebuilt point-cloud map, while the terrain, Bonxai voxel-mapping, graph-mapping, and exploration modules update traversability, occupancy, frontiers, and the cluster decision graph online.
Panoramic images are segmented with Mask2Former, fused with LiDAR measurements, and accumulated as the semantic scene consumed by the object-search supervisor.
The supervisor dispatches \texttt{EXPLORE}, \texttt{RELOCATE}, and \texttt{NAV} goals through the same \texttt{TeamMission} action interface used by the navigation stack.
In this deployment it runs at 1~Hz, permits an LLM relocation decision every 5~s, accepts object clusters containing at least 10 semantic voxels, and confirms a target after three supporting observations.

\begin{figure}[t]
	\centering
	\includegraphics[width=\columnwidth]{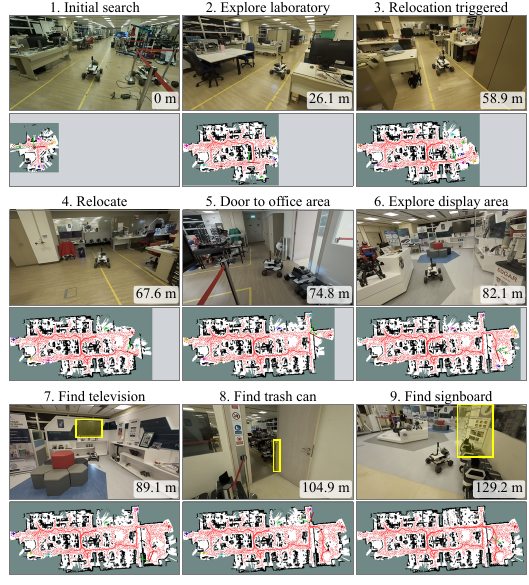}
	\caption{Representative events from a continuous real-world ObjectNav run in which the next request was issued after each successful target confirmation.
		Each labeled pair shows the first-person observation (top) and accumulated map, graph, and trajectory (bottom).
		The sequence highlights the relocation trigger, passage through the office doorway, and terminal observations for television, trash can, and signboard; all map panels use the final viewport.
		The reported distance is cumulative robot travel from the beginning of the sequence, rather than geodesic distance to a target.}
	\label{fig:real_world_search_sequence}
\end{figure}

To demonstrate continuous ObjectNav, a new natural-language target was issued only after the supervisor had successfully confirmed the preceding target.
The requested sequence was television $\rightarrow$ trash can $\rightarrow$ signboard, and the mapping and navigation stack was not reset between requests.
Figure~\ref{fig:real_world_search_sequence} aligns the robot's first-person observations with the accumulated RViz map at representative decision and completion events.

The first six pairs expose the mechanism behind the physical search rather than only its final detections.
The map grows from a small local component into a connected representation of the workspace; at pair~3 (58.9~m), contextual mismatch opens the relocation gate, after which the robot reaches the office doorway at 74.8~m and extends the route into the display area.
Pairs~7--9 show the terminal views for three distinct requests: the television is found at 89.1~m, followed by the trash can after another 15.8~m and the signboard after a further 24.3~m.
The later map panels change less than the early ones because subsequent searches reuse already accumulated occupancy, topology, and semantic evidence, while their trajectories terminate at different target locations.
Together, the sequence demonstrates online cross-region relocation, persistent-map reuse across requests, and target-conditioned navigation on the physical ROS stack.

\section{Discussion and Conclusion}

SOR-Nav achieves the strongest reported SR and SPL in our three-dataset comparison, and the HM3D-v2 ablation shows complementary gains from context-gated cross-region relocation and TSP-based frontier ordering.
Together with the physical sequence, these results support the central design claim: ObjectNav benefits from separating the decision to leave the current search context from the subsequent choice of a reachable destination.

Cross-paper numbers remain sensitive to success distance, episode filtering, and multi-floor handling; our results describe the complete unfiltered evaluation rather than controlled reimplementations of every baseline.
The nested ablation also does not isolate every representation component, and the physical result is a qualitative deployment rather than a statistical benchmark.
Future work will evaluate the system under matched protocols and across more physical environments.
Overall, SOR-Nav provides a structured and deployable coupling between an autonomous semantic exploration system and a context-gated, LLM-driven object-search supervisor.

\bibliographystyle{IEEEtran}
\bibliography{refs}

@article{sysnav,
  title        = {SysNav: Multi-Level Systematic Cooperation Enables Real-World, Cross-Embodiment Object Navigation},
  author       = {Zhu, Haokun and Li, Zongtai and Liu, Zihan and Guo, Kevin and Lin, Zhengzhi and Cai, Yuxin and Chen, Guofei and Lv, Chen and Wang, Wenshan and Oh, Jean and Zhang, Ji},
  journal      = {arXiv preprint arXiv:2603.06914},
  year         = {2026}
}

@article{vlfm,
  title        = {VLFM: Vision-Language Frontier Maps for Zero-Shot Semantic Navigation},
  author       = {Yokoyama, Naoki and Ha, Sehoon and Batra, Dhruv and Wang, Jiuguang and Bucher, Bernadette},
  journal      = {arXiv preprint arXiv:2312.03275},
  year         = {2023}
}

@article{l3mvn,
  title        = {L3MVN: Leveraging Large Language Models for Visual Target Navigation},
  author       = {Yu, Bangguo and Kasaei, Hamidreza and Cao, Ming},
  journal      = {arXiv preprint arXiv:2304.05501},
  year         = {2023}
}

@article{esc,
  title        = {ESC: Exploration with Soft Commonsense Constraints for Zero-shot Object Navigation},
  author       = {Zhou, Kaiwen and Zheng, Kaizhi and Pryor, Connor and Shen, Yilin and Jin, Hongxia and Getoor, Lise and Wang, Xin Eric},
  journal      = {arXiv preprint arXiv:2301.13166},
  year         = {2023}
}

@article{voronav,
  title        = {VoroNav: Voronoi-based Zero-shot Object Navigation with Large Language Model},
  author       = {Wu, Pengying and Mu, Yao and Wu, Bingxian and Hou, Yi and Ma, Ji and Zhang, Shanghang and Liu, Chang},
  journal      = {arXiv preprint arXiv:2401.02695},
  year         = {2024}
}

@article{sgnav,
  title        = {SG-Nav: Online 3D Scene Graph Prompting for LLM-based Zero-shot Object Navigation},
  author       = {Yin, Hang and Xu, Xiuwei and Wu, Zhenyu and Jie, Zhou and Lu, Jiwen},
  journal      = {arXiv preprint arXiv:2410.08189},
  year         = {2024}
}

@article{openfmnav,
  title        = {OpenFMNav: Towards Open-Set Zero-Shot Object Navigation via Vision-Language Foundation Models},
  author       = {Kuang, Yuxuan and Lin, Hai and Jiang, Meng},
  journal      = {arXiv preprint arXiv:2402.10670},
  year         = {2024}
}

@article{trihelper,
  title        = {TriHelper: Zero-Shot Object Navigation with Dynamic Assistance},
  author       = {Zhang, Lingfeng and Zhang, Qiang and Wang, Hao and Xiao, Erjia and Jiang, Zixuan and Chen, Honglei and Xu, Renjing},
  journal      = {arXiv preprint arXiv:2403.15223},
  year         = {2024}
}

@article{instructnav,
  title        = {InstructNav: Zero-shot System for Generic Instruction Navigation in Unexplored Environment},
  author       = {Long, Yuxing and Cai, Wenzhe and Wang, Hongcheng and Zhan, Guanqi and Dong, Hao},
  journal      = {arXiv preprint arXiv:2406.04882},
  year         = {2024}
}

@article{apexnav,
  title        = {ApexNav: An Adaptive Exploration Strategy for Zero-Shot Object Navigation with Target-centric Semantic Fusion},
  author       = {Zhang, Mingjie and Du, Yuheng and Wu, Chengkai and Zhou, Jinni and Qi, Zhenchao and Ma, Jun and Zhou, Boyu},
  journal      = {arXiv preprint arXiv:2504.14478},
  year         = {2025}
}

@article{intentnav,
  title        = {IntentNav: Learning Spatial-Visual Object Navigation from Human Demonstrations},
  author       = {Cai, Yuxin and Li, Zongtai and Wang, Maonan and Bao, Muyi and Zhu, Haokun and Bai, Ruofei and Zhao, Ding and Li, Zirui and Wang, Wenshan and Yau, Wei-Yun and Zhang, Ji and Lv, Chen},
  journal      = {arXiv preprint arXiv:2606.08029},
  year         = {2026}
}

@article{ddppo,
  title        = {{DD-PPO}: Learning Near-Perfect PointGoal Navigators from 2.5 Billion Frames},
  author       = {Wijmans, Erik and Kadian, Abhishek and Morcos, Ari and Lee, Stefan and Essa, Irfan and Parikh, Devi and Savva, Manolis and Batra, Dhruv},
  journal      = {arXiv preprint arXiv:1911.00357},
  year         = {2019}
}

@inproceedings{habitatweb,
  title        = {Habitat-Web: Learning Embodied Object-Search Strategies from Human Demonstrations at Scale},
  author       = {Ramrakhya, Ram and Undersander, Eric and Batra, Dhruv and Das, Abhishek},
  booktitle    = {Proceedings of the IEEE/CVF Conference on Computer Vision and Pattern Recognition},
  pages        = {5173--5183},
  year         = {2022}
}

@inproceedings{pirlnav,
  title        = {{PIRLNav}: Pretraining with Imitation and {RL} Finetuning for ObjectNav},
  author       = {Ramrakhya, Ram and Batra, Dhruv and Wijmans, Erik and Das, Abhishek},
  booktitle    = {Proceedings of the IEEE/CVF Conference on Computer Vision and Pattern Recognition},
  pages        = {17896--17906},
  year         = {2023}
}

@inproceedings{semexp,
  title        = {Object Goal Navigation Using Goal-Oriented Semantic Exploration},
  author       = {Chaplot, Devendra Singh and Gandhi, Dhiraj and Gupta, Abhinav and Salakhutdinov, Ruslan},
  booktitle    = {Advances in Neural Information Processing Systems},
  volume       = {33},
  pages        = {4247--4258},
  year         = {2020}
}

@inproceedings{poni,
  title        = {Potential Functions for ObjectGoal Navigation with Interaction-Free Learning},
  author       = {Ramakrishnan, Santhosh Kumar and Chaplot, Devendra Singh and Al-Halah, Ziad and Malik, Jitendra and Grauman, Kristen},
  booktitle    = {Proceedings of the IEEE/CVF Conference on Computer Vision and Pattern Recognition},
  pages        = {18890--18900},
  year         = {2022}
}

\end{document}